\documentclass[letterpaper]{article} 
\usepackage{aaai24}  
\usepackage{times}  
\usepackage{helvet}  
\usepackage{courier}  
\usepackage[hyphens]{url}  
\usepackage{graphicx} 
\usepackage{natbib}  
\usepackage{caption} 
\usepackage{amsmath}
\usepackage{amsthm}
\usepackage{booktabs}
\usepackage[ruled,linesnumbered,vlined]{algorithm2e}
\SetKwInOut{Input}{Input}
\SetKwInOut{Output}{Output}
\usepackage{algorithmic}
\usepackage{colortbl}
\usepackage{enumitem}
\setlist{leftmargin=*}
\usepackage{amsfonts}
\usepackage{xcolor}
\usepackage{subcaption}
\usepackage{mwe}
\usepackage{multirow}
\usepackage{mdframed}
\usepackage{bbding}
\usepackage[capitalize]{cleveref}
\usepackage{threeparttable}
\DeclareMathOperator{\tr}{tr}

\newcommand{\first}[1]{\textbf{#1}}
\newcommand{\second}[1]{\underline{#1}}
\newcommand{\third}[1]{#1}

\title{Rethinking Graph Masked Autoencoders through\\Alignment and Uniformity}
\author{
    Liang Wang\textsuperscript{\rm 1,2}\equalcontrib, 
    Xiang Tao\textsuperscript{\rm 1,2}\equalcontrib, 
    Qiang Liu\textsuperscript{\rm 1,2}, 
    Shu Wu\textsuperscript{\rm 1,2}\thanks{Corresponding author.}, 
    Liang Wang\textsuperscript{\rm 1,2}
}
\affiliations{
    \textsuperscript{\rm 1}Center for Research on Intelligent Perception and Computing\\ State Key Laboratory of Multimodal Artificial Intelligence Systems\\ Institute of Automation, Chinese Academy of Sciences\\
    \textsuperscript{\rm 2}School of Artificial Intelligence, University of Chinese Academy of Sciences\\
    \{liang.wang, xiang.tao\}@cripac.ia.ac.cn, \{qiang.liu, shu.wu, wangliang\}@nlpr.ia.ac.cn
}

\usepackage{bibentry}

\newcommand{\themodel}{\textsf{AUG-MAE}\xspace}

\begin{document}

\maketitle

\begin{abstract}
Self-supervised learning on graphs can be bifurcated into contrastive and generative methods. 
Contrastive methods, also known as graph contrastive learning (GCL), have dominated graph self-supervised learning in the past few years, but the recent advent of graph masked autoencoder (GraphMAE) rekindles the momentum behind generative methods.
Despite the empirical success of GraphMAE, there is still a dearth of theoretical understanding regarding its efficacy. Moreover, while both generative and contrastive methods have been shown to be effective, their connections and differences have yet to be thoroughly investigated.
Therefore, we theoretically build a bridge between GraphMAE and GCL, and prove that the node-level reconstruction objective in GraphMAE implicitly performs context-level GCL. Based on our theoretical analysis, we further identify the limitations of the GraphMAE from the perspectives of alignment and uniformity, which have been considered as two key properties of high-quality representations in GCL. 
We point out that GraphMAE's alignment performance is restricted by the masking strategy, and the uniformity is not strictly guaranteed. 
To remedy the aforementioned limitations, we propose an \underline{A}lignment-\underline{U}niformity enhanced \underline{G}raph \underline{M}asked \underline{A}uto\underline{E}ncoder, named \themodel.
Specifically, we propose an easy-to-hard adversarial masking strategy to provide hard-to-align samples, which improves the alignment performance. Meanwhile, we introduce an explicit uniformity regularizer to ensure the uniformity of the learned representations.
Experimental results on benchmark datasets demonstrate the superiority of our model over existing state-of-the-art methods.
The code is available at: \url{https://github.com/AzureLeon1/AUG-MAE}.
\end{abstract}

\section{1. Introduction}

Graph self-supervised learning can be categorized into two distinct types, contrastive and generative methods~\cite{survey2021tkde,survey2022tpami,survey2022tkde}. 
Motivated by the InfoMax principle, contrastive methods, also known as graph contrastive learning (GCL), maximize the mutual information between positive pairs. The contrastive loss is proved to asymptotically optimize two properties, representation \textit{alignment} and \textit{uniformity}, which are considered to lead to high-quality representations~\cite{au}.
On the other hand, the basic idea behind generative methods is to reconstruct the masked portions of data with generative models, such as autoencoders and autoregressive models. The reconstruction process reveals inherent data patterns and encode them into learned representations. 

In the past few years, contrastive methods have dominated graph self-supervised learning due to their superior performance, and have gained sufficient theoretical analysis and understanding.
Recently, graph masked autoencoder (GraphMAE)~\cite{GraphMAE} is proposed and demonstrates that generative methods can also achieve competitive, and even better, performance when appropriately designed. 
GraphMAE analyzes the deficiencies of early generative methods in terms of reconstruction target, decoder structure, and optimization objective. This model addresses these deficiencies in a sophisticated manner and achieves performance beyond that of the contrastive methods.
Many subsequent studies take GraphMAE as a foundation to further improve the model structure and apply it to different domains~\cite{GMAE, MaskGAE, S2GAE, AutoCF, MAERec}.
These studies spark renewed interest and reflection on generative methods.

However, despite the recent empirical success of GraphMAE, there is still a lack of sufficient understanding regarding its efficacy. 
Additionally, it remains unknown whether there exists a connection between GraphMAE and GCL.
Specifically, the following questions arise:
\textit{\textbf{Why is GraphMAE effective? Are GraphMAE and GCL completely different methods, or do they share any commonality?}}

To answer these questions, we conducted a theoretical analysis of GraphMAE. To facilitate the understanding of the relationship between GraphMAE and GCL, we do not analyze GraphMAE independently, but build a bridge between GraphMAE and GCL. 
Specifically, we first view the learning process of GraphMAE as using the contexts (ego-graphs) of the masked nodes to restore the original features of these nodes. Then, we theoretically prove that the node-level reconstruction loss in GraphMAE is lower bounded by the context-level alignment loss. This indicates that GraphMAE has the ability to align positive pairs defined in contrastive learning, and \textit{GraphMAE implicitly performs context-level GCL} in its learning process.

\begin{figure}
    \centering
    \begin{subfigure}[b]{0.27\linewidth}
        \centering
        \includegraphics[width=\linewidth]{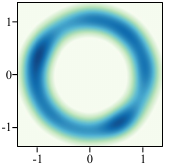}
        \caption{GCL}
        \label{fig:pilot-1}
    \end{subfigure}
    \hspace{15pt}
    \begin{subfigure}[b]{0.27\linewidth}
        \centering
        \includegraphics[width=\linewidth]{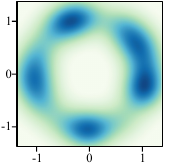}
        \caption{GraphMAE}
        \label{fig:pilot-2}
    \end{subfigure}
    \caption{Distribution of nodes representations on the unit hypersphere learned by GCL (taking GRACE~\cite{GRACE} as an example) and GraphMAE~\cite{GraphMAE}. The representations learned by GCL is more uniformly distributed than GraphMAE.}
    \label{fig:pilot}
\end{figure}

Since we have established the connection between GraphMAE and GCL through theoretical analysis, we are able to identify the limitations of GraphMAE from the perspective of representation alignment and uniformity:
(1) For alignment, although GraphMAE is proven to have the ability to align representations of positive pairs, the practical alignment performance not only depends on the model's ability, but also influenced by the masking strategy. Further, the random masking strategy adopted in GraphMAE ignores the difficulty of aligning positive samples. (2) For uniformity, the representation uniformity is not strictly guaranteed in GraphMAE. Specifically, GraphMAE can naturally avoid full feature collapse, i.e., the learned representations do not collapse to be the same. However, as shown in~\cref{fig:pilot}, we observe that the representations still suffer from partial dimensional collapse~\cite{DirectCLR, ContraNorm}, i.e., the representations shrink along a certain dimension and are not uniformly distributed in the feature space. Therefore, the uniformity of the representations can also be improved.

To overcome these limitations, we propose an \underline{A}lignment-\underline{U}niformity enhanced \underline{G}raph \underline{M}asked \underline{A}uto\underline{E}ncoder, named \themodel.
Specifically, we propose an easy-to-hard adversarial masking strategy to provide richer hard-to-align samples, which improves the alignment performance. Moreover, we introduce an explicit uniformity regularizer to ensure the uniformity of the learned representations.
Experimental results on benchmark datasets demonstrate the superiority of our model over existing state-of-the-art methods.
Meanwhile, the learned representations are better aligned and more uniformly distributed in the feature space. 
The main contributions of our work are outlined as follows:
\begin{itemize}
    \item We conduct a theoretical analysis of GraphMAE and demonstrate that it implicitly performs context-level GCL. Further, we identify the limitations of GraphMAE from the perspective of alignment and uniformity. 
    \item We propose an \themodel model. For alignment, we employ an easy-to-hard adversarial masking strategy to generate hard-to-align positive pairs. For uniformity, we introduce an explicit uniformity regularizer.
    \item We conduct extensive experiments on benchmark datasets, which show that \themodel outperforms state-of-the-art methods on downstream tasks, and achieves better alignment and uniformity.
\end{itemize}
\section{2. Related Work}

In this section, we succinctly review existing studies for graph self-supervised learning and two measurements of representation learning, i.e., alignment and uniformity.

\noindent\textbf{Graph Self-Supervised Learning.}
Graph self-supervised learning has been proposed as a promising paradigm for learning graph representations without labels.
Existing methods can be categorized into contrastive and generative. 

Contrastive methods learn meaningful representations by maximizing the mutual information between augmented views~\cite{CREAME,STENCIL}. 
Some early efforts focus on contrastive modes~\cite{PyGCL,GCA,GCC,GraphCL},
and several recent studies focus on the negative-sample-free technique~\cite{BGRL}.
Despite the progress made by these studies, GCL still relies on intricate designs.

Although early generative methods lagged behind contrastive methods, the recently proposed GraphMAE~\cite{GraphMAE} has greatly improved the empirical performance of generative methods through sophisticated designs and triggered many subsequent studies, such as WGDN~\cite{WGDN}, SeeGera~\cite{SeeGera}, and HGMAE~\cite{HGMAE}. However, the theoretical support of GraphMAE is still not thoroughly investigated. 

\noindent\textbf{Alignment and Uniformity.}
Several studies analyze how the contrastive objective influences the representation learning. \citet{au} first identify two properties induced from contrastive learning objective: alignment of positive pairs and uniformity of the representation distribution on the hypersphere.
Both alignment and uniformity play vital roles in enhancing the discriminative and generalization abilities of contrastive learning models, then these two properties are utilized to measure and improve the quality of learned representations~\cite{SimGRACE,QRec,AURL,U-MAE}.
\section{3. Preliminary}
\subsection{3.1. Problem Formulation}
Let $\mathcal{G} = (\mathcal{V}, \mathbf{A}, \mathbf{X})$ denote a given graph, where $\mathcal{V}=\{v_i\}_{i=1}^N$ represents the node set. The adjacency matrix and the feature matrix are denoted as $\mathbf{A}\in\{0,1\}^{N\times N}$ and $\mathbf{X}\in\mathbb{R}^{N\times d}$ respectively, where $\boldsymbol{x}_i\in\mathbb{R}^d$ is the feature of $v_i$ and $\mathbf{A}_{ij}=1$ iff there is an edge between $v_i$ and $v_j$. In the setting of graph self-supervised learning, there is no available label information during training. Our goal is to learn a GNN encoder $f(\cdot)$ receiving the graph structure and features, and producing low-dimensional node representations.
We denote $\mathbf{Z}=f(\mathbf{X}, \mathbf{A})\in\mathbb{R}^{N\times d^{\prime}}$ as the learned node representations, where $\boldsymbol{z}_i\in\mathbb{R}^{d^{\prime}}$ is the representation of node $v_i$. 
The representations are $l_2$-normalized on the unit hypersphere $\mathcal{S}^{d^{\prime}-1}$, which is common in machine learning.

\subsection{3.2. Graph Masked Autoencoders}
We choose the canonical GraphMAE~\cite{GraphMAE} as the object of analysis because it serves as the foundation for various subsequent models.
GraphMAE first randomly sample a subset of nodes $\widetilde{\mathcal{V}} \in \mathcal{V}$ based on a uniform distribution.
Then, the node features of these selected nodes are masked:
\begin{equation}
\tilde{\boldsymbol{x}}_i=\left\{\begin{array}{ll}
\boldsymbol{x}_{[\textrm{MASK}]} & v_i \in \widetilde{\mathcal{V}}, \\
\boldsymbol{x}_i & v_i \notin \widetilde{\mathcal{V}},
\end{array}\right.
\end{equation}
where $\boldsymbol{x}_{[\textrm{MASK}]}\in \mathbb{R}^d$ is the learnable mask token, and $\tilde{\boldsymbol{x}}_i\in\mathbb{R}^d$ is the feature of node $v_i$ after masking.

The GraphMAE model $h=g \circ f$ is an encoder-decoder architecture, where a GNN-based encoder 
$f$ maps the contexts (ego-graphs) of masked nodes to latent features, and a GNN-based decoder $g$ reconstructs the features of masked nodes from the latent contexts. 
The task performed by GraphMAE can be interpreted as the reconstruction of the original features of masked nodes from their $(l_e+l_d)$-hop contexts, where $l_e$ and $l_d$ denote the numbers of encoder layers and decoder layers, respectively.
We use $c_i$ to denote the $(l_e+l_d)$-hop context of node $v_i$ after masking.
The reconstructed feature of node $v_i$ is $\hat{\boldsymbol{x}}_i = h(c_i) = g(f(c_i))$.

Finally, GraphMAE adopts the scaled cosine error (SCE) on masked features as the reconstruction loss:
\begin{equation}
\begin{aligned}
    \mathcal{L}_{\mathrm{SCE}} = \mathbb{E}_{v_i \in \widetilde{\mathcal{V}}}\left(1- {\boldsymbol{x}_i}^{\top} h(c_i)\right)^\gamma, 
\end{aligned}
\label{sce}
\end{equation}
where the cosine similarity is simply represented as the dot product since the original feature and reconstructed feature are both $l_2$-normalized. The scaling factor $\gamma \geq 1$ is a hyper-parameter that adjusts the weight of each sample with the reconstruction error.

\subsection{3.3. Alignment and Uniformity Loss}
The alignment and uniformity properties are necessary for high-quality representations, and highly related to the contrastive learning~\cite{au}. 

Alignment loss aims to make the representations of semantically similar samples as close as possible, and thus the representations can be invariant to unneeded noise factors. Alignment loss is consistent with the contrastive objective of maximizing the agreement of positive pairs. Formally, the alignment loss is defined as:
\begin{equation}
\mathcal{L}_{\text {Align}} = {\mathbb{E}}_{\left(v, v^{+}\right)\sim p_{\mathrm{pos}}}\left\|\boldsymbol{z}-\boldsymbol{z}^{+}\right\|^2.
\label{eq:align_loss}
\end{equation}
where $p_{\mathrm{pos}}$ is the distribution of positive pairs, and $\boldsymbol{z}$ is the learned representations of data sample $v$.

Uniformity loss prefers the uniform distribution on the unit hypersphere, so as to preserves maximal information of data. 
Uniformity helps to avoid feature collapse and learn discriminable representations.
The uniformity loss is defined as the logarithm of the average pairwise Gaussian potential:
\begin{equation}
\mathcal{L}_{\text {Uni}} = \log {\mathbb{E}}_{ v_i, v_j \stackrel{\text { i.i.d. }}{\sim} p_{\mathrm{data}} } e^{-t\left\|\boldsymbol{z}_i-\boldsymbol{z}_j\right\|^2},
\label{eq:uni_loss}
\end{equation}
where $p_{\mathrm{data}}$ is the distribution of data, and $t$ is a hyperparameter for Gaussian potential kernel. In the contrastive objective, uniformity is achieved by pulling away the distance between negative pairs.
\section{4. Alignment-Uniformity Enhanced Graph Masked Autoencoders}

In this section, we first conduct a theoretical analysis of GraphMAE and identify its limitations from the perspective of alignment and uniformity. Subsequently, we propose our \themodel model to overcome these limitations.

\subsection{4.1. Theoretical Understanding of GraphMAE}

There is a viewpoint that generative and contrastive methods adhere to different philosophies, where contrastive methods deal with the inter-data information and generative methods focus on the intra-data information~\cite{survey2021tkde}. 
However, we perform a deep analysis and give an insight that \textit{generative methods, such as GraphMAE, perform implicit context-level graph contrastive learning}.

\begin{figure*}[htb]
\centering
\includegraphics[width=0.98\linewidth]{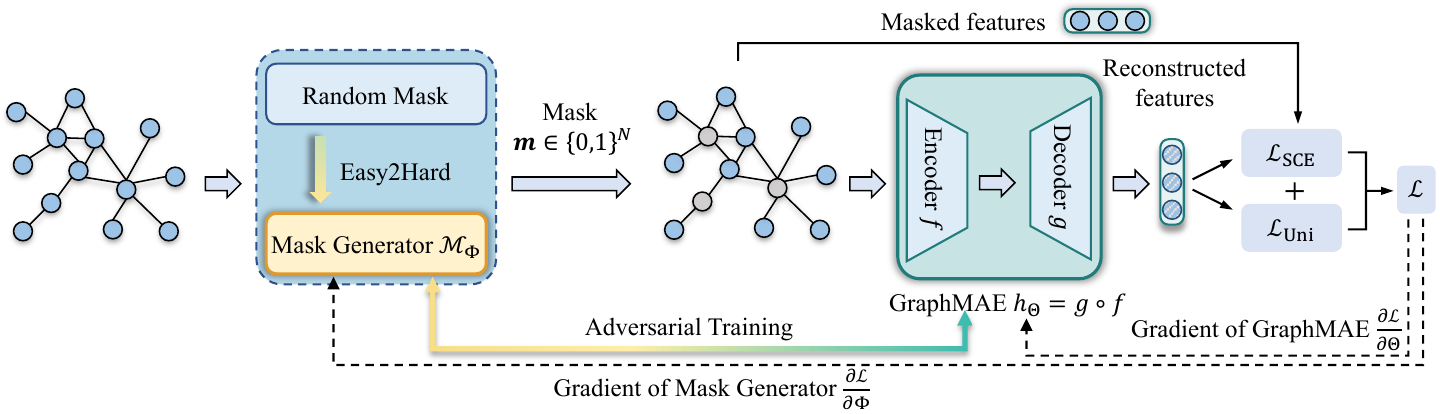}
\caption{The overall framework of our proposed \themodel model. We propose an easy-to-hard adversarial masking strategy to provide hard-to-align positive pairs, so as to improve the alignment ability of GraphMAE. Additionally, we introduce an explicit uniformity regularizer $\mathcal{L}_{\mathrm{Uni}}$ into the objective to enhance the uniformity of learned representations.}
\label{overview}
\end{figure*}

Since GarphMAE is based on the autoencoder framework, we first assume that it is capable of accomplishing the vanilla autoencoding task: reconstructing the original input.

\begin{mdframed}[backgroundcolor=gray!15,hidealllines=true,innerleftmargin=2pt,innerrightmargin=2pt]
\noindent\textbf{Assumption 4.1.\label{assum4-1}} For any graph decoder $g$, we assume that there exists a pseudo-inverse graph encoder $f_g$ such that the resulting pseudo graph autoencoder $h_g = g \circ f_g$ satisfies $\mathbb{E}_{v_i \in \widetilde{\mathcal{V}}}\left\|h_g(\boldsymbol{x})-\boldsymbol{x}\right\|^2 \leq \varepsilon$.
\end{mdframed}
This assumption is valid, since the GNN-based encoder and decoder degenerate to MLPs when input contains only one node, and MLPs have been proven to be universal approximators of arbitrary continuous functions~\cite{universal-1}.

\begin{mdframed}[backgroundcolor=gray!15,hidealllines=true,innerleftmargin=2pt,innerrightmargin=2pt]
\noindent\textbf{Theorem 4.2.\label{theo4-2}} Under Assumption 4.1, the SCE loss in \cref{sce} can be lower bounded by a pretext loss:
\begin{equation}
\begin{aligned}
&\mathcal{L}_{\text {SCE }}(h) \geq \frac{\gamma}{2}\mathcal{L}_{\text {Pretext }}(h)-\frac{\gamma}{2}\varepsilon+\text { const }, \\
\end{aligned}
\label{eq1}
\end{equation}
where $\mathcal{L}_{\text {Pretext }}(h) =-{\mathbb{E}}_{v_i \in \widetilde{\mathcal{V}}} h_g\left(\boldsymbol{x}_i\right)^{\top} h\left(c_i\right)$.
\end{mdframed}

Please refer to Appendix for the detailed proof of Theorem 4.2.
Then we define the context-level alignment loss, which is an objective of context-level GCL, and we prove it is a lower bound of pretext loss above.
\begin{mdframed}[backgroundcolor=gray!15,hidealllines=true,innerleftmargin=2pt,innerrightmargin=2pt]
\noindent\textbf{Definition 4.3. \textit{(Context-Level Alignment Loss)}\label{def4-3}} The alignment loss for positive context pairs $(c, c^+)$ is defined as:
\begin{equation}
\mathcal{L}_{\text {Align }}^{\text{c}}(h)=-{\mathbb{E}}_{{(c, c^{+})\sim p_{\mathrm{pos}}^c }} h\left(c\right)^{\top} h\left(c^{+}\right).
\label{eq2}
\end{equation}
\end{mdframed}

This loss in the form of dot product and the loss in the form of mean squared error in \cref{eq:align_loss} are equivalent because the reconstructed features $\{h(c)\}_{c\sim p_{\mathrm{data}}^c }$ are all normalized.

\begin{mdframed}[backgroundcolor=gray!15,hidealllines=true,innerleftmargin=2pt,innerrightmargin=2pt]
\noindent\textbf{Theorem 4.4.\label{theo4-4}} The pretext loss in \cref{eq1} can be lower bounded by the context-level alignment loss in \cref{eq2}:
\begin{equation}
\mathcal{L}_{\text {Pretext }}(h) \geq \frac{1}{2} \mathcal{L}_{\text {Align}}^{\text{c}}(h)+\text{const}.
\end{equation}
\end{mdframed}

The detailed proof of Theorem 4.4 can be found in Appendix.
Combining Theorem 4.2 and Theorem 4.4, we arrive at the main theorem showing that GraphMAE’s node-level reconstruction loss can be bounded by the alignment loss of the positive context pairs drawn from the masked nodes.

\begin{mdframed}[backgroundcolor=gray!15,hidealllines=true,innerleftmargin=2pt,innerrightmargin=2pt]
\noindent\textbf{Theorem 4.5.\label{theo4-5}} Under Assumption 4.1, GraphMAE’s nodel-level reconstruction loss in \cref{sce} can be lower bounded by the context-level alignment loss in \cref{eq2}:
\begin{equation}
\begin{aligned}
   &\mathcal{L}_{\text{SCE}}(h) \geq \frac{\gamma}{4} \mathcal{L}_{\text {Align}}^{\text{c}}(h)-\frac{\gamma}{2}\varepsilon+\text{const}\\
\end{aligned}
\label{conclustion}
\end{equation}
\end{mdframed}

Following Theorem 4.5, a small GraphMAE's reconstruction loss implies a small context-level alignment loss, which indicates that GraphMAE implicitly aligns the representations of positive context pairs.

\vspace{4pt}
\noindent\textbf{Intuitive explanation.} 
Here, we provide an intuitive explanation for our theoretical result. We reiterate that GraphMAE employs the autoencoder to reconstruct the masked node features based on the contexts of these mask nodes. When two (or more) masked nodes have the same or very similar features, then their contexts have the same reconstruction goal, and thus these contexts will be encoded as similar latent representations. In the paradigm of contrastive learning, these contexts can be considered as positive pairs.

\subsection{4.2. Limitations of GraphMAE\label{subsection:limitation}}
The theoretical analysis above builds a bridge between GraphMAE and GCL. 
Therefore, we can identify the limitations of GraphMAE with the well-established tools in GCL. Specifically, we further discuss the limitations of GraphMAE from the perspective of alignment and uniformity.

For alignment, although GraphMAE is proven to have the ability to align positive pairs, the practical alignment effect not only depends on the model's ability, but also \textit{influenced by the masking strategy}. Concretely,
the optimization objective is computed as an expectation over a distribution, which is essentially decided by the distribution of the mask.
Further, the uniform random masking strategy adopted by GraphMAE ignores the imbalanced distribution of easy and hard samples, thus cannot provide enough information about hard-to-align pairs. 

For uniformity, the representation uniformity is \textit{not strictly guaranteed in GraphMAE}. 
Specifically, GraphMAE avoids full feature collapse, i.e., the learned representations do not collapse to a fixed point in the feature space, as long as the masked features are not identical. However, reconstruction loss cannot lead to uniformly distributed representations. In \cref{fig:pilot}, we observe that the representations still suffer from partial dimensional collapse.
Therefore, the representation uniformity can also be improved.
\subsection{4.3. The proposed AUG-MAE Model}

To overcome the aforementioned limitations, we propose \themodel. The overall framework is illustrated in \cref{overview}.

\subsubsection{Adversarial Masking}
GraphMAE randomly selects nodes to mask based on the uniform distribution, which ignores the imbalanced distribution of easy and hard samples. Random masking is able to sample a large number of easy-to-align positive paris, but yields only a few hard-to-align positive pairs. Therefore, GraphMAE does not work well on these hard samples, limiting the quality of the learned representations. For this reason, we propose an adversarial masking strategy to mine more hard-to-align positive pairs.

To generate mask adaptively, we devise a GNN-based mask generator $\mathcal{M}$ with parameter $\Phi$. Given a graph $\mathcal{G}$, the mask generator produces a probability vector of adversarial masking $prob_{\mathrm{adv}} \in [0,1]^N$:
\begin{equation}
    prob_{\mathrm{adv}} = \mathcal{M}_{\Phi}(\mathcal{G}),
\label{eq:adv_mask}
\end{equation}
where $prob_{\mathrm{adv},i}$ denotes the probability of node $v_i$ being masked. Then, the Gumbel-Softmax~\cite{Gumbel-Softmax} is applied on each $prob_{\mathrm{adv},i}$ to generate a differentiable binary mask vector $\boldsymbol{m}\in \{0,1\}^N$:
\begin{equation}
    \boldsymbol{m}_i = \sigma(\frac{1}{\tau}(\log(\frac{prob_{\mathrm{adv},i}}{1-prob_{\mathrm{adv},i}}+(\epsilon_0-\epsilon_1)))),
\label{eq:adv_mask2}
\end{equation}
where $\epsilon_0,\epsilon_1$ are Gumbel random noises sampled from $\text{Gumbel}(0,1)$, $\tau$ is a temperature hyperparameter, and $\sigma$ is the sigmoid function.
We use the $\widetilde{V}_{\mathrm{adv}} = \{v_i|\boldsymbol{m}_i=1, i=1,2,\cdots,N\}$ to represent the set of masked nodes. 
Then, the node features are masked with generated mask:
\begin{equation}
\tilde{\boldsymbol{x}}_i=\left\{\begin{array}{ll}
\boldsymbol{x}_{[\mathrm{MASK}]} & v_i \in \widetilde{V}_{\mathrm{adv}}, \\
\boldsymbol{x}_i & v_i \notin \widetilde{V}_{\mathrm{adv}}.
\end{array}\right.
\end{equation}

In previous studies~\cite{MAE,GraphMAE}, mask raito has been empirically proven to be important for masked autoencoding.
However, \cref{eq:adv_mask} and (\ref{eq:adv_mask2}) cannot control the ratio of masked nodes.
To adjust the mask ratio of the mask generator, we introduce a ratio regularizer in the form of $1/sin(\cdot)$ to its optimization objective. Formally, the parameters of the mask generator $\Phi$ can be updated by optimizing:
\begin{equation}
\Phi^{\star}=\arg \max _{\Phi} (\mathcal{L}_{\mathrm{SCE}}(\mathcal{G};\Theta, \Phi) - \lambda_1\sin (\frac{\pi}{N}\sum_{i=1}^N \boldsymbol{m}_i)^{-1}),
\label{eq:adv1}
\end{equation}
where $\frac{1}{N}\sum_{i=1}^{N}\boldsymbol{m}_i$ is mask ratio. $\lambda_1$ is the weight of the ratio regularizer, which controls the mask ratio. The large $\lambda_1$ encourages a mask ratio close to 50\%.
Adversarially, the parameters of GraphMAE $\Theta$ can be learned by:
\begin{equation}
\Theta^{\star}=\arg \min _{\Theta} \mathcal{L}_{\mathrm{SCE}}(\mathcal{G};\Theta, \Phi).
\label{eq:adv2}
\end{equation}
During the adversarial training process, $\Theta$ and $\Phi$ are iteratively optimized so that the mask generator and GraphMAE evolve simultaneously. The mask generator gradually learns to generate hard-to-align positive pairs, while GraphMAE gradually learns how to align these pairs.

\subsubsection{Easy-to-Hard Training}
To ensure the training stability, we adopt an easy-to-hard strategy for training. In the early stage of training, we employ the random masking to generate the mask, utilizing plenty of easy samples to initialize the model parameters. The advantage of this initialization is that the model can initially have the ability to handle easy samples and have a relatively accurate judgment of the difficulty of the samples. 
During the training process, we gradually increase the weight of the adversarial masking and decrease the weight of the random masking, so that the model can obtain further improvement from the hard samples in the later stage of training. The easy-to-hard process is controlled by:
\begin{equation}
    prob(t) = (1-\alpha_{\mathrm{adv}}(t))\cdot prob_{\mathrm{rand}} + \alpha_{\mathrm{adv}}(t)\cdot prob_{\mathrm{adv}}(t),
    \label{eq:prob}
\end{equation}
where $t$ denotes the current epoch, $T$ denotes the total epochs, and $prob\in [0,1]^N$ denotes the masking probability vector, which is the weighted sum of the masking probability vector of random masking $prob_{\mathrm{rand}}$ and that of adversarial masking $prob_{\mathrm{adv}}$. Then the mask $\boldsymbol{m}$ is sampled from $prob$.

During the training process (from epoch $0$ to epoch $T$), the weight of the adversarial mask $\alpha_{\mathrm{adv}}$ grows from  $\alpha_0$ to $\alpha_T$. Correspondingly, the weight of the random mask decreases from (1-$\alpha_0$) to (1-$\alpha_T$). The change of $\alpha_{\mathrm{adv}}$ is defined as:
\begin{equation}
    \alpha_{\mathrm{adv}}(t) = \alpha_0+\Delta\alpha(t) =\alpha_0+ (\frac{t}{T})^{\eta} \cdot(\alpha_T-\alpha_0),
    \label{eq:adv_weight}
\end{equation}
where $\alpha_0, \alpha_T \in [0, 1], \alpha_0< \alpha_T$. $\eta$ controls the rate of weight growth. 
$\eta=1$ indicates a linear growth from easy to hard, and $\eta \neq 1$ indicates a non-linear growth.

\begin{table*}[t]
    \centering
    \begin{threeparttable}
    \resizebox{1\linewidth}{!}{
    \begin{tabular}{c|c|ccccccccc|>{\columncolor{gray!20}}c}
    \toprule
        ~ & Method & Cora & CiteSeer & PubMed & Ogbn-arxiv & PPI & Reddit & Corafull & Flickr & WikiCS & A.R. \\ 
        \midrule
        \multirow{6}*{Contrastive} & DGI & 82.3 $\pm$ 0.6 & 71.8 $\pm$ 0.7 & 76.8 $\pm$ 0.6 & 70.3 $\pm$ 0.2 & 63.8 $\pm$ 0.2 & 94.0 $\pm$ 0.1 & 48.2 $\pm$ 0.5 & 45.0 $\pm$ 0.2 & 64.8 $\pm$ 0.6 & 7.89\\ 
        ~ & MVGRL & 83.5 $\pm$ 0.4 & \second{73.3 $\pm$ 0.5} & 80.1 $\pm$ 0.7 & - & - & - & 52.6 $\pm$ 0.5 & - & 64.8 $\pm$ 0.7 & 5.20\\ 
        ~ & GRACE & 81.9 $\pm$ 0.4 & 71.2 $\pm$ 0.5 & 80.6 $\pm$ 0.4 & \third{71.5 $\pm$ 0.1} & 69.7 $\pm$ 0.2 & 94.7 $\pm$ 0.1 & 45.2 $\pm$ 0.1 & - & \third{68.0 $\pm$ 0.7} & 6.50\\ 
        ~ & BGRL & 82.7 $\pm$ 0.6 & 71.1 $\pm$ 0.8 & 79.6 $\pm$ 0.5 & \second{71.6 $\pm$ 0.1} & 73.6 $\pm$ 0.2 & 94.2 $\pm$ 0.1 & 47.4 $\pm$ 0.5 & 39.4 $\pm$ 0.1 & 65.5 $\pm$ 1.5 & 6.56\\ 
        ~ & InfoGCL & 83.5 $\pm$ 0.3 & \first{73.5 $\pm$ 0.4} & 79.1 $\pm$ 0.2 & - & - & - & - & - & - & 4.67\\ 
        ~ & CCA-SSG & \second{84.0 $\pm$ 0.4} & 73.1 $\pm$ 0.3 & \third{81.0 $\pm$ 0.4} & 71.2 $\pm$ 0.2 & 73.3 $\pm$0.2 & 95.1 $\pm$ 0.1 & \second{53.5 $\pm$ 0.4} & 49.1 $\pm$ 0.1 & 67.4 $\pm$0.9 & 3.89\\
        \midrule
        \multirow{5}*{Generative} & SeeGera & 82.8 $\pm$ 0.3 & 71.6 $\pm$ 0.2 & 79.2 $\pm$ 0.3 & 71.2 $\pm$ 0.3 & 73.4 $\pm$ 0.3 & 95.2 $\pm$ 0.2 & 52.0 $\pm$ 0.4 & \third{49.4 $\pm$ 0.5} & 65.8 $\pm$ 0.2 & 5.78\\ 
        ~ & MaskGAE & 82.6 $\pm$ 0.3 & 73.1 $\pm$ 0.6 & \second{81.0 $\pm$ 0.3} & 71.2 $\pm$ 0.3 & \third{73.9 $\pm$ 0.3} & \third{95.4 $\pm$ 0.1} & 52.2 $\pm$ 0.1 & 49.1 $\pm$ 0.4 & 66.0 $\pm$ 0.2 & 4.78\\ 
        ~ & GraphMAE & \third{84.0 $\pm$ 0.6} & 73.1 $\pm$ 0.4 & 80.9 $\pm$ 0.4 & 71.3 $\pm$ 0.6 & \second{74.1 $\pm$ 0.4} & \second{95.8 $\pm$ 0.4} & \third{53.3 $\pm$ 0.4} & \second{49.5 $\pm$ 0.5} & \second{70.6 $\pm$ 0.9} & \second{3.00}\\ 
        \cmidrule{2-12}
        ~ & AUG-MAE & \first{84.3 $\pm$ 0.4} & \third{73.2 $\pm$ 0.4} & \first{81.4 $\pm$ 0.4} & \first{71.9 $\pm$ 0.2} & \first{74.3 $\pm$ 0.1} & \first{96.1 $\pm$ 0.1} & \first{57.6 $\pm$ 0.3} & \first{50.3 $\pm$ 0.2} & \first{71.7 $\pm$ 0.6} & \first{1.22}\\ 
    \bottomrule
    \end{tabular}}
    \end{threeparttable}
    \caption{Node classification results on benchmarks. We report Micro-F1(\%) score for PPI and accuracy(\%) for the other datasets. The best results are highlighted in \textbf{bold} and the runner ups are highlighted with \underline{underlines}. A.R. means the average rank.}
    \label{tab:node}
\end{table*}

\begin{table*}[t]
    \centering
    \begin{threeparttable}
    \resizebox{0.82\linewidth}{!}{
    \begin{tabular}{c|c|cccccc|>{\columncolor{gray!20}}c}
    \toprule
        ~ & Method & IMDB-B & IMDB-M & PROTEINS & COLLAB & MUTAG & REDDIT-B & A.R. \\ 
        \midrule
        \multirow{7}*{Contrastive} & Graph2vec & 71.10 $\pm$ 0.54 & 50.44 $\pm$ 0.87 & 73.30 $\pm$ 2.05 & - & 83.15 $\pm$ 9.25 & 75.78 $\pm$ 1.03 & 7.00\\ 
        ~ & InfoGraph & 73.03 $\pm$ 0.87 & 49.69 $\pm$ 0.53 & 74.44 $\pm$ 0.31 & 70.65 $\pm$ 1.13 & \third{89.01 $\pm$ 1.13} & 82.50 $\pm$ 1.42 & 5.17\\ 
        ~ & GraphCL & 71.14 $\pm$ 0.44 & 48.58 $\pm$ 0.67 & 74.39 $\pm$ 0.45 & 71.36 $\pm$ 1.15 & 86.80 $\pm$ 1.34 & \second{89.53 $\pm$ 0.84} & 5.83\\ 
        ~ & JOAO & 70.21 $\pm$ 3.08 & 49.20 $\pm$ 0.77 & \third{74.55 $\pm$ 0.41} & 69.50 $\pm$ 0.36 & 87.35 $\pm$ 1.02 & 85.29 $\pm$ 1.35 & 6.33\\ 
        ~ & GCC & 72.0 & 49.4 & - & 78.9 & - & \first{89.8} & 4.50\\ 
        ~ & MVGRL & 74.20 $\pm$ 0.70 & 51.20 $\pm$ 0.50 & - & - & \second{89.70 $\pm$ 1.10} & 84.50 $\pm$ 0.60 & 4.00\\ 
        ~ & InfoGCL & \third{75.10 $\pm$ 0.90} & \second{51.40 $\pm$ 0.80} & - & \third{80.00 $\pm$ 1.30} & \first{91.20 $\pm$ 1.30} & - & \second{2.25}\\ 
        \midrule
        \multirow{3}*{Generative} & GraphMAE & \second{75.30 $\pm$ 0.59} & \third{51.35 $\pm$ 0.78} & \second{75.30 $\pm$ 0.52} & \second{80.32 $\pm$ 0.42} & 88.19 $\pm$ 1.26 & 87.83 $\pm$ 0.25 & 3.00\\ 
        \cmidrule{2-9}
        ~ & AUG-MAE & \first{75.56 $\pm$ 0.61} & \first{51.80 $\pm$ 0.86} & \first{75.83 $\pm$ 0.24} & \first{80.48 $\pm$ 0.50} & 88.28 $\pm$ 0.98 & \third{87.98 $\pm$ 0.43} & \first{1.83}\\ 
    \bottomrule
    \end{tabular}
    }
    \end{threeparttable}
    \caption{Graph classification results on benchmarks. We report accuracy(\%) for all datasets.  The best results are highlighted in \textbf{bold} and the runner ups are highlighted with \underline{underlines}. A.R. means the average rank.}
    \label{tab:graph}
\end{table*}

\begin{table*}[t]
    \centering
    \begin{threeparttable}
    \begin{minipage}{\textwidth}
    \centering
    \resizebox{0.85\linewidth}{!}{
    \begin{tabular}{c|ccc|cccccc}
    \toprule
        \multirow{2}*{Model} & \multicolumn{3}{c|}{Component}  & \multicolumn{6}{c}{Node Classification Dataset} \\ 
        \cmidrule{2-10}
        ~ & Adv & E2H & Uni & Cora & CiteSeer & PubMed & Ogbn-arxiv & PPI & Reddit \\ 
        \midrule
        GraphMAE &- &- &-  & 84.04 $\pm$ 0.58 & 73.11 $\pm$ 0.38 & 80.94 $\pm$ 0.47 & 71.32 $\pm$ 0.55 & 74.09 $\pm$ 0.37 & 95.79 $\pm$ 0.36 \\ 
        Variant 1 & {\color{black}\Checkmark} &- &- & 84.30 $\pm$ 0.75 & 73.18 $\pm$ 0.59 & 81.25 $\pm$ 0.53 & 71.43 $\pm$ 0.07 & 74.12 $\pm$ 0.36 & 95.97 $\pm$ 0.36 \\ 
        Variant 2 &- &- & {\color{black}\Checkmark}  & 84.24 $\pm$ 0.58 & 73.16 $\pm$ 0.54 & 81.18 $\pm$ 0.44 & 71.42 $\pm$ 0.30 & 74.13 $\pm$ 0.20 & 95.90 $\pm$ 0.15 \\ 
        Variant 3 & {\color{black}\Checkmark} & {\color{black}\Checkmark} &-  & 84.20 $\pm$ 0.54 & 73.14 $\pm$ 0.58 & 81.28 $\pm$ 0.43 & 71.50 $\pm$ 0.30 & 74.24 $\pm$ 0.05 & 95.85 $\pm$ 0.25 \\ 
        Variant 4 & {\color{black}\Checkmark} &- & {\color{black}\Checkmark}  & \textbf{84.32 $\pm$ 0.45} & 73.18 $\pm$ 0.48 & 81.30 $\pm$ 0.54 & 71.50 $\pm$ 0.15 & 74.13 $\pm$ 0.36 & 96.00 $\pm$ 0.10 \\ 
        AUG-MAE & {\color{black}\Checkmark} & {\color{black}\Checkmark} & {\color{black}\Checkmark}  & 84.30 $\pm$ 0.38 & \textbf{73.20 $\pm$ 0.44} & \textbf{81.35 $\pm$ 0.44} & \textbf{71.86 $\pm$ 0.22} & \textbf{74.30 $\pm$ 0.11} & \textbf{96.07 $\pm$ 0.03} \\ 
        \bottomrule
    \end{tabular}
    }
    \end{minipage}

    \begin{minipage}{\textwidth}
    \centering
    \resizebox{0.85\linewidth}{!}{
    \begin{tabular}{c|ccc|cccccc}
    \toprule
        \multirow{2}*{Model} &\multicolumn{3}{c|}{Component}  & \multicolumn{6}{c}{Graph Classification Dataset} \\ 
        \cmidrule{2-10}
        ~ & Adv & E2H & Uni & IMDB-B & IMDB-M & PROTEINS & COLLAB & MUTAG & REDDIT-B \\ 
        \midrule
        GraphMAE &- &- &- & 75.30 $\pm$ 0.59 & 51.35 $\pm$ 0.78 & 75.30 $\pm$ 0.52 & 80.32 $\pm$ 0.42 & 88.19 $\pm$ 1.26 & 87.83 $\pm$ 0.25 \\ 
        Variant 1 & {\color{black}\Checkmark} &- &- & 74.32 $\pm$ 0.75 & 49.92 $\pm$ 0.92 & 75.72 $\pm$ 1.04 & 79.88 $\pm$ 0.69 & 87.00 $\pm$ 1.40 & 87.35 $\pm$ 0.35 \\ 
        Variant 2 &- &- & {\color{black}\Checkmark} & 75.40 $\pm$ 0.60 & 51.50 $\pm$ 0.49 & 75.50 $\pm$ 0.51 & 80.37 $\pm$ 0.44 & 88.20 $\pm$ 1.56 & 87.90 $\pm$ 0.42 \\ 
        Variant 3 & {\color{black}\Checkmark} & {\color{black}\Checkmark} &- & 75.20 $\pm$ 0.88 & 51.59 $\pm$ 1.36 & 75.65 $\pm$ 0.53 & 80.26 $\pm$ 0.44 & 88.01 $\pm$ 1.11 & 87.90 $\pm$ 0.27 \\ 
        Variant 4 & {\color{black}\Checkmark} &- & {\color{black}\Checkmark} &74.56 $\pm$ 0.58 & 50.22 $\pm$ 0.69 & 75.75 $\pm$ 0.43 & 80.07 $\pm$ 0.45 & 87.41 $\pm$ 1.27 & 87.47 $\pm$ 0.45 \\ 
        AUG-MAE & {\color{black}\Checkmark} & {\color{black}\Checkmark} & {\color{black}\Checkmark} & \textbf{75.56 $\pm$ 0.61} & \textbf{51.80 $\pm$ 0.86} & \textbf{75.83 $\pm$ 0.24} & \textbf{80.48 $\pm$ 0.50} & \textbf{88.28 $\pm$ 0.98} & \textbf{87.98 $\pm$ 0.43} \\ 
        \bottomrule
    \end{tabular}
    }
    \end{minipage}
    \begin{tablenotes}    
        \footnotesize
        \item \qquad\quad * Adv: adversarial masking, E2H: easy-to-hard training, Uni: uniformity regularizer.
      \end{tablenotes} 
    \end{threeparttable}
    \caption{Ablation analysis, in which we keep different components in our \themodel to form variants. We report accuracy(\%) of these variants for node and graph classification datasets. The best performance is highlighted in \textbf{bold}.}
    \label{tab:ablation}
\end{table*}

\begin{figure*}
    \centering
    \includegraphics[width=0.94\linewidth]{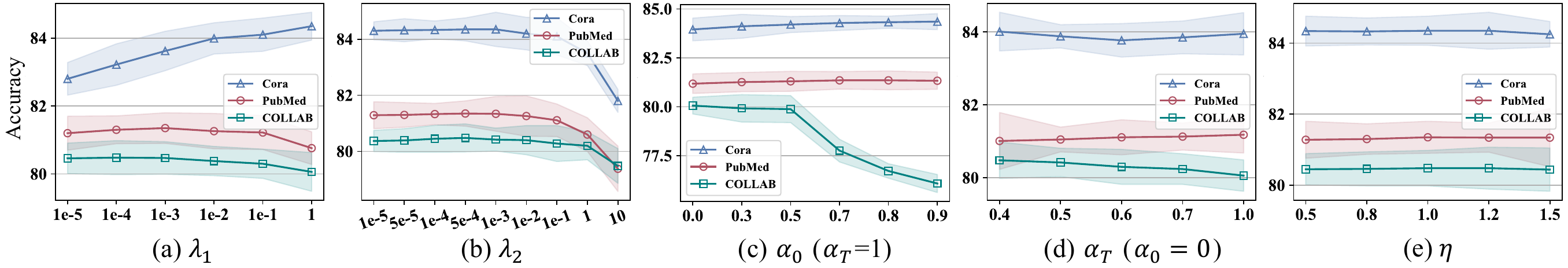}
       \caption{Effect of different hyper-parameters. The y-axis represents accuracy(\%).}
       \label{fig:sen}
\end{figure*}

\subsubsection{Explicit Uniformity Regularizer}
As mentioned in theoretical analysis, GraphMAE naturally avoids full feature collapse, but still suffers from partial dimensional collapse. Therefore, we explicitly introduce a uniformity regularizer into the objective of GraphMAE. The objective defined by \cref{eq:adv2} is updated as:
\begin{equation}
\Theta^{\star}=\arg \min _{\Theta} (\mathcal{L}_{\mathrm{SCE}}(\mathcal{G};\Theta, \Phi) + (1-\alpha_{\mathrm{adv}})\lambda_2 \mathcal{L}_{\mathrm{Uni}}(\mathcal{G};\Theta)),
\label{eq:adv3}
\end{equation}
where $\mathcal{L}_{\mathrm{Uni}}$ denotes the uniformity regularizer given in \cref{eq:uni_loss}, and $\lambda_2$ is the weight of uniformity regularizer.
It is worth noting that we desire that the representations of all nodes are uniformly distributed on the hypersphere, and not just the representations of hard samples. Therefore, uniformity regularization should be paired with the random masking strategy. To achieve it, we also use $(1-\alpha_{\mathrm{adv}})$ to control the impact of regularizer during the training process.

To help better understand the adversarial training process, we provide the brief pseudo-code of it in Appendix.
\section{5. Experiments}

In this section, we conduct experiments to evaluate the effectiveness of \themodel. We analyze it by answering the following questions:
\textbf{RQ1}: How does \themodel perform compared with graph self-supervised learning baselines, especially GraphMAE, in various downstream tasks?
\textbf{RQ2}: How does each component influence the performance of \themodel?
\textbf{RQ3}: How do key hayper-parameters influence the performance of \themodel?
\textbf{RQ4}: How does the alignment and uniformity of the representations learned by \themodel, compare with GCL and GraphMAE?

\subsection{5.1. Evaluation Setups}

\noindent\textbf{Datasets.}
We select nine node classification datasets (i.e., Cora, Citeseer~\cite{cora_citeseer}, Pubmed~\cite{pubmed}, Ogbn-arxiv~\cite{ogb}, PPI, Reddit, Corafull~\cite{corafull}, Flickr~\cite{flickr}, and WikiCS~\cite{WikiCS}), and six graph classification datasets (i.e., IMDB-B, IMDB-M, PROTEINS, COLLAB, MUTAG, and REDDIT-B~\cite{TUDataset}). Standard data splits are adopted.

\noindent\textbf{Baselines.}
We consider both contrastive methods and generative methods as baselines. Node-level GCL baselines are compared in the node classification task, including DGI~\cite{DGI}, MVGRL~\cite{MVGRL}, GRACE, BGRL~\cite{BGRL}, InfoGCL~\cite{InfoGCL}, and CCA-SSG~\cite{CCA-SSG}.
In graph classification task, compared graph-level GCL baselines are Graph2vec~\cite{Graph2vec}, InfoGraph~\cite{InfoGraph}, GraphCL, JOAO~\cite{JOAO}, GCC, MVGRL, and InfoGCL.
For generative methods, we select SeeGera, MaskGAE, and GraphMAE as baselines.

Detailed evaluation setups can be found in Appendix.

\subsection{5.2. Performance Comparison (RQ1)}

We compare \themodel with the baselines and the results are summarized in \cref{tab:node} and \cref{tab:graph}.
On both node classification and graph classification tasks, \themodel outperforms all graph self-supervised baselines on most datasets.

Among the baseline models, GraphMAE, as a recently proposed generative method, achieves competitive performance with state-of-the-art contrastive methods.
Since our work focuses on analyzing the limitations of GraphMAE and improving it, we first focus on comparing our method with GraphMAE. Our method outperforms GraphMAE on all datasets for both downstream tasks. This verifies the feasibility of improving GraphMAE from the perspective of alignment and uniformity, as well as validates the effectiveness of our proposed strategies.

On the downstream node classification task, the representations learned by our \themodel are able to achieve the highest accuracy on all datasets except CiteSeer.
On the graph classification task, the representations learned by our \themodel also have the highest accuracy on most datasets. 
However, on the MUTAG and REDDIT-B datasets, although our method outperforms GraphMAE, it still does not outperform some GCL methods.
We speculate that the reason may be that generative methods focus more on context-level information, while node-level and graph-level information are also important on these datasets.

\subsection{5.3. Ablation Study (RQ2)}

To analyze the effectiveness of the different components, we conduct an ablation study. 
The results are summarized in \cref{tab:ablation}. We have the following observations from this table.

\noindent\textbf{Effect of Adversarial Masking.}
By comparing GraphMAE and Variant 1 (also Variant 2 and Variant 4), we can observe that the adversarial masking is better than random masking, and steadily boosts the performance on node classification.
On the graph classification task, the straightforward introduction of adversarial masking does not seem to be helpful. But when combined with the easy-to-hard training strategy, Adv+E2H can effectively improve the performance. This can be observed by comparing Variant 2 and \themodel.

\noindent\textbf{Effect of Easy-to-Hard Training.}
The easy-to-hard strategy is designed to assist in the adversarial masking.
From the difference between the results of Variant 1 and Variant 3 (also Variant 4 and \themodel) on graph classification datasets, we can find that the easy-to-hard training is very important on graph-level representation learning. 

\noindent\textbf{Effect of Uniformity Regularizer.}
By comparing GraphMAE and Variant 2 (also Variant 3 and \themodel), we can observe that uniformity regularizer effectively improves the performance of both node classification and graph classification, which verifies the effectiveness of this regularizer.

\subsection{5.4. Sensitivity Analysis (RQ3)}

\cref{fig:sen} shows the effect of varied hyper-parameter values, from which we have the following observations.

\noindent\textbf{Effect of weight of ratio regularizer $\lambda_1$.}
This weight affects the result of representation learning by affecting the mask ratio.
As shown in \cref{fig:sen}(a), when $\lambda_1 \geq 1$, the mask ratio is around 0.5 and the best performance is achieved on Cora. When $\lambda_1=1e-3$, the mask ratio is around 0.75 and the best performance is achieved on PubMed and COLLAB. They are consistent with the optimal mask ratios for random masking provided by GraphMAE~\cite{GraphMAE}.

\noindent\textbf{Effect of weight of uniformity regularizer $\lambda_2$.}
The optimal choice for this weight on most datasets is 5e-4 or 5e-5. We find that uniformity regularizer impairs accuracy when the weight is larger than 1e-1. 
This is because the excessive pursuit of uniformity can damage the distinguishability of learned representations.

\noindent\textbf{Effect of parameters controlling easy-to-hard $\alpha_0, \alpha_T, \eta$.}
From \cref{fig:sen}(c) and \cref{fig:sen}(d) we observe that the impact of $\alpha_0$ and $\alpha_T$ on graph classification datasets is obvious.
Taking COLLAB as an example, appropriate values of $\alpha_0$ and $\alpha_T$ can effectively improve the performance.
We tune $\eta$ in the range of $[0.5,1.5]$, and find that our model is not sensitive to the change of $\eta$. Relatively speaking, the best performance is achieved with $\eta=1$ on most of the datasets.

\begin{figure}
    \centering
    \begin{subfigure}[b]{0.34\linewidth}
        \centering
        \includegraphics[width=\linewidth]{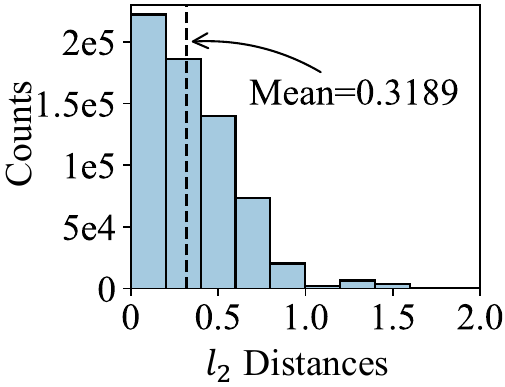}
        \caption{GCL}
    \end{subfigure}
    \begin{subfigure}[b]{0.319\linewidth}
        \centering
        \includegraphics[width=\linewidth]{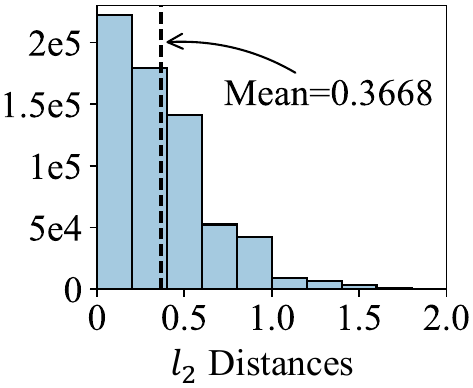}
        \caption{GraphMAE}
    \end{subfigure}
    \begin{subfigure}[b]{0.319\linewidth}
        \centering
        \includegraphics[width=\linewidth]{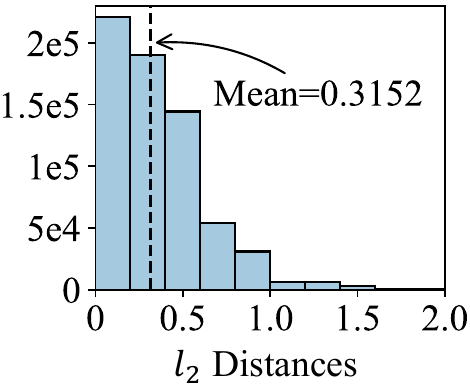}
        \caption{AUG-MAE}
    \end{subfigure}
       \caption{$l_2$ distances between positive representations of Cora learned by GCL, GraphMAE, and AUG-MAE. The smaller mean distance indicates the better alignment.}
       \label{fig:distance}
\end{figure}

\subsection{5.5. Alignment and Uniformity Analysis (RQ4)}
On the Cora dataset, we take nodes with the same label as positive samples, and compute $l_2$ distance between them (also called supervised alignment loss~\cite{DirectAU}). The statistical results are shown in \cref{fig:distance}, in which we also plot the mean distances with dashed lines. Since smaller mean distance indicates better alignment, we can observe that \themodel can align similar samples better than GraphMAE, and even slightly better than GCL.

\begin{figure}
    \centering
    \begin{subfigure}[b]{0.26\linewidth}
        \centering
        \includegraphics[width=\linewidth]{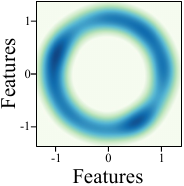}
        \caption{GCL}
    \end{subfigure}
    \hspace{6pt}
    \begin{subfigure}[b]{0.26\linewidth}
        \centering
        \includegraphics[width=\linewidth]{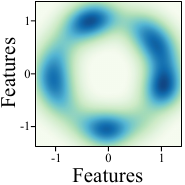}
        \caption{GraphMAE}
    \end{subfigure}
    \hspace{6pt}
    \begin{subfigure}[b]{0.26\linewidth}
        \centering
        \includegraphics[width=\linewidth]{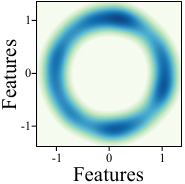}
        \caption{AUG-MAE}
    \end{subfigure}
    \caption{Representation distributions of Cora on $\mathcal{S}^1$ learned by GCL, GraphMAE, and AUG-MAE. We plot distributions with Gaussian kernel density estimation in $\mathbb{R}^2$.}
    \label{fig:vis}
\end{figure}

We also visualize the representation distributions learned by GCL, GraphMAE, and our \themodel in \cref{fig:vis}.
Compared with GraphMAE, representations learned by \themodel achieve better uniformity, i.e., representations are more uniformly distributed on $\mathcal{S}^1$.
\section{6. Conclusion}

We theoretically prove that the node-level reconstruction in GraphMAE implicitly performs context-level GCL. Based on this, we identify the limitations of GraphMAE from the perspective of alignment and uniformity.
To overcome them, we propose \themodel equipped with an easy-to-hard adversarial masking strategy and an explicit uniformity regularizer.
Experimental results show that \themodel produces representations with better alignment and uniformity, and surpasses baselines on downstream tasks.

\section{Acknowledgments}
This work is jointly sponsored by National Natural Science Foundation of China (62206291, 62141608).

\bibliography{paper}

\clearpage
\appendix
\section{Appendix}

The organization of the appendix is as follows:
\begin{itemize}
    \item Appendix A: Proofs of Theorems;
    \item Appendix B: Pseudo-code of Training Process;
    \item Appendix C: Details of Datasets;
    \item Appendix D: Details of Baselines;
    \item Appendix E: Experimental Settings;
    \item Appendix F: Implementation Details.
\end{itemize}

\section{A. Proofs of Theorems}

\subsection{A.1. Proof of Theorem 4.2}

With Assumption 4.1, and considering that the features are all normalized, we have:
\begin{align*}
    &\mathcal{L}_{\mathrm{SCE}}= \underset{v_i \in \widetilde{\mathcal{V}}}{\mathbb{E}}\left(1- {\boldsymbol{x}_i^{\top}} h(c_i)\right)^\gamma \\
    &\geq \underset{v_i \in \widetilde{\mathcal{V}}}{\mathbb{E}} (1 - \gamma{\boldsymbol{x}_i^{\top}} h(c_i)) \tag{Bernoulli's inequality} \\
    &= \underset{v_i \in \widetilde{\mathcal{V}}}{\mathbb{E}} (1 - \gamma(1 - \frac{1}{2} \|\boldsymbol{x}_i - h(c_i)\|^2)) \tag{features are normalized} \\
    &= 1-\gamma + \frac{\gamma }{2}\underset{v_i \in \widetilde{\mathcal{V}}}{\mathbb{E}} \|\boldsymbol{x}_i - h(c_i)\|^2 \\
    &= 1-\gamma + \frac{\gamma }{2}\underset{v_i \in \widetilde{\mathcal{V}}}{\mathbb{E}} (\|\boldsymbol{x}_i - h(c_i)\|^2 + \varepsilon) -  \frac{\gamma }{2}\varepsilon \tag{Assumption 4.1}\\
    &\geq 1-\gamma + \frac{\gamma }{2}\underset{v_i \in \widetilde{\mathcal{V}}}{\mathbb{E}} (\|\boldsymbol{x}_i - h(c_i)\|^2 + \|h_g(\boldsymbol{x}_i)-\boldsymbol{x}_i\|^2) -\frac{\gamma }{2}\varepsilon.
\end{align*}

Since the inequality $\|a+b\|^2 \leq 2(\|a\|^2+\|b\|^2)$ holds, we can further obtain:
\begin{align*}
    \mathcal{L}_{\mathrm{SCE}} &\geq 1-\gamma + \frac{\gamma }{4}\underset{v_i \in \widetilde{\mathcal{V}}}{\mathbb{E}} \|h_g(\boldsymbol{x}_i) - h(c_i)\|^2 - \frac{\gamma }{2}\varepsilon \\
    &= 1-\gamma + \frac{\gamma }{4}\underset{v_i \in \widetilde{\mathcal{V}}}{\mathbb{E}} (2-2h_g\left(\boldsymbol{x}_i\right)^{\top} h\left(c_i\right))) - \frac{\gamma }{2}\varepsilon \\
    &= -\frac{\gamma }{2} \underset{{v_i \in \widetilde{\mathcal{V}}}}{\mathbb{E}} h_g\left(\boldsymbol{x}_i\right)^{\top} h\left(c_i\right) -\frac{\gamma }{2}\varepsilon + 1-\frac{\gamma }{2} \\
    &= \frac{\gamma}{2}\mathcal{L}_{\text {Pretext }}(h)-\frac{\gamma}{2}\varepsilon+\text { const }.
\end{align*}

\begin{table*}[t]
\centering
\resizebox{0.95\textwidth}{!}{
\begin{tabular}{lccccccccc}
\toprule
Datasets      & Cora  & Citeseer & Pubmed & Ogbn-arxiv & PPI     & Reddit     & Corafull & Flickr  & WikiCS  \\
\midrule
\#Nodes       & 2,708 & 3,327    & 19,717 & 169,343    & 56,944  & 232,965    & 19,793   & 89,250  & 11,701  \\
\#Edges       & 5,278 & 4,732    & 44,338 & 1,166,243  & 818,736 & 11,606,919 & 126,842  & 899,756 & 431,726 \\
\#Classes     & 7     & 6        & 3      & 40         & 121     & 41         & 70       & 7       & 10     \\
\#Features    & 1,433 & 3,703    & 500    & 767        & 50      & 602        & 8,710    & 500     & 300   \\
\bottomrule
\end{tabular}}
\caption{Statistics of node classification datasets.}
\label{tab:node-data}
\end{table*}

\begin{table*}[t]
\centering
\resizebox{0.75\textwidth}{!}{
\begin{tabular}{lcccccccc}
\toprule
Datasets      & IMDB-B  & IMDB-M & PROTEINS & COLLAB   & MUTAG & REDDIT-B  \\ \midrule
\#Graphs      & 1,000   & 1,500  & 1,113    & 5,000    & 188   & 2,000     \\
\#Classes     & 2       & 3      & 2        & 3        & 2     & 2         \\
Avg. \#Nodes  & 19.8    & 13.0   & 39.1     & 74.5     & 17.9  & 429.7     \\ 
Avg. \#Edges  & 96.5    & 65.9   & 72.8     & 2,457.5  & 19.8  & 497.8     \\ 
\bottomrule
\end{tabular}}
\caption{Statistics of graph classification datasets.}
\label{tab:graph-data}
\end{table*}

\subsection{A.2. Proof of Theorem 4.4}

\begin{figure}
    \centering
    \includegraphics[width=0.8\linewidth]{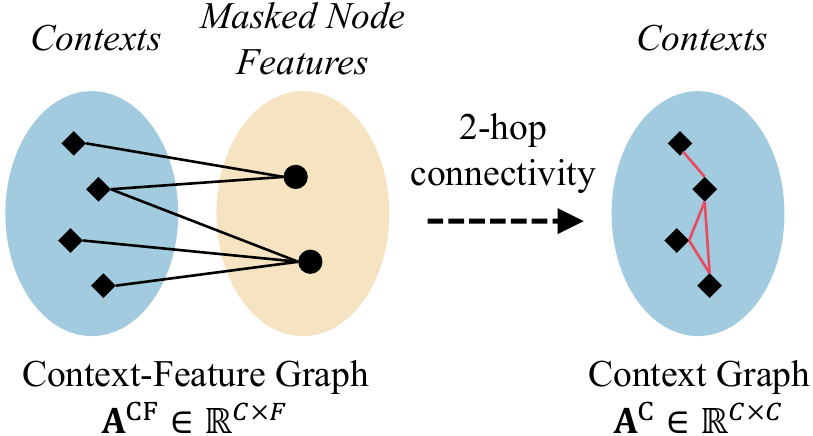}
    \caption{Context graph based on 2-hop connectivity in context-feature graph.}
    \label{fig:context_graph}
\end{figure}

To facilitate the proof of Theorem 4.4, we define the context-feature graph induced from the mask, as illustrated in \cref{fig:context_graph}. The context-feature graph is a bipartite graph that indicates the relations between masked node features and their contexts, which are utilized to reconstruct the masked node features. 

We use $\mathcal{F}=\{f_i\}_{i=1}^{F}$ to denote the set of node features, and $\mathcal{C}=\{c_i\}_{i=1}^C$ to denote the set of all possible contexts.
For any node $v_i$ that is masked, there exists a feature node $f_j$ whose feature is equal to the feature of the masked node, i.e., $\boldsymbol{x}_i=f_j$.
Note that nodes with the same features correspond to the same feature node in our graph.
The adjacency matrix of context-feature graph is defined as $\mathbf{A}_{\mathrm{CF}}\in \mathbb{R}^{C\times F}$, where $(\mathbf{A}_{\mathrm{CF}})_{i,j}=w_{c_i,f_j}$ between $c_i \in \mathcal{C}, f_j \in \mathcal{F}$ is defined as their joint probability. In other words, there is an edge between $c_i$ and $f_j$ if and only if they belong to a complementary context-feature pair. The normalized adjacency matrix is defined as $\tilde{\mathbf{A}}_{\mathrm{CF}} = \mathbf{D}_{\mathrm{C}}^{-1/2}\mathbf{A}_{\mathrm{CF}}\mathbf{D}_{\mathrm{F}}^{-1/2}$, where $\mathbf{D}_{\mathrm{C}}, \mathbf{D}_{\mathrm{F}}$ are the diagonal degree matrices with elements $d_{c_i}=\sum_{f_j}w_{c_i,f_j}$ and $d_{f_j}=\sum_{c_i}w_{c_i,f_j}$.

With our defined context-feature graph, the $\mathcal{L}_{\text {Pretext }}(h)$ in \cref{eq1} can be reformulated as:
\begin{equation}
    \mathcal{L}_{\text {Pretext }}(h) =-\underset{{v_i \in \widetilde{\mathcal{V}}}}{\mathbb{E}} h_g\left(\boldsymbol{x}_i\right)^{\top} h\left(c_i\right) = -\tr(\mathbf{H}_g^{\top} \tilde{\mathbf{A}}_{\mathrm{CF}} \mathbf{H})
\end{equation}
where $\mathbf{H}$ denotes the output matrix of $h$ on $\mathcal{C}$ whose $c_i$-th row is $\mathbf{H}_{c_i} = \sqrt{d_{c_i}}h(c_i)$, and $\mathbf{H}_g$ denotes the output matrix of $h_g$ on $\mathcal{F}$ whose $f_j$-th row is $(\mathbf{H}_g)_{f_j} = \sqrt{d_{f_j}}h_g(f_j)$.
Then, we have:
\begin{align*}
&\mathcal{L}_{\text {Pretext }}(h)  =-\tr(\mathbf{H}_g^{\top} \tilde{\mathbf{A}}_{\mathrm{CF}} \mathbf{H}) \\
& \geq-\frac{1}{2}\left(\left\|\mathbf{H}_g\right\|^2_\mathrm{F}+\left\| \tilde{\mathbf{A}}_{\mathrm{CF}}\mathbf{H}\right\|_\mathrm{F}^2\right) \tag{$\tr(\mathbf{AB}) \leq \frac{1}{2}\left(\|\mathbf{A}\|_\mathrm{F}^2+\|\mathbf{B}\|_\mathrm{F}^2\right)$}\\
& =-\frac{1}{2} \operatorname{tr}\left( \tilde{\mathbf{A}}_{\mathrm{CF}}^{\top} \tilde{\mathbf{A}}_{\mathrm{CF}}  \mathbf{H} \mathbf{H}^{\top}\right)-\frac{1}{2} \tag{$\left\|\mathbf{H}_g\right\|_\mathrm{F}^2=\sum_{f_j} d_{f_j}\left\|h_g\left(f_j\right)\right\|^2=1$} \\
& =-\frac{1}{2} \sum_{c, c^{+}} \sum_{f_j} \frac{w_{c,f_j}w_{c^{+},f_j}}{d_{f_j}} h\left(c\right)^{\top} h\left(c^{+}\right)-\frac{1}{2} \\
& =-\frac{1}{2} \sum_{c, c^{+}} (\mathbf{A}_{\mathrm{C}})_{c, c^{+}} h\left(c\right)^{\top} h\left(c^{+}\right)-\frac{1}{2} \\
& =\frac{1}{2} \mathcal{L}_{\text {align }}^{\text{c}}(h)-\frac{1}{2},
\end{align*}
where $\|\mathbf{H}\|_\mathrm{F}$ denotes the Frobenius norm of $\mathbf{H}$. Here we define $(\mathbf{A}_{\mathrm{C}})_{c, c^{+}}=\sum_{f_j} \frac{w_{c,f_j}w_{c^{+},f_j}}{d_{f_j}}$, which represents the joint probability of positive context pair $(c, c^+)$. As shown in \cref{fig:context_graph}, we construct the homogeneous context graph based on the 2-hop connectivity from the context-node bipartite graph. The edge weights represent the probability of forming positive pairs between contexts. In this case, $\mathbf{A}_{\mathrm{C}} \in \mathbb{R}^{C\times C}$ represents the adjacency matrix of this context graph.

\section{B. Pseudo-code of Training Process}
To help better understand the adversarial training process, we provide the brief pseudo-code of it in Algorithm~\ref{algo}.
\begin{algorithm}[h]
\caption{Adversarial training process of \themodel}
\label{algo}
\While{not converge}{
    // Update parameters of mask generator $\Phi$ \\
    Compute masking probability $prob$ according to \cref{eq:prob} and (\ref{eq:adv_weight}), and generate mask $\boldsymbol{m}$ based on $prob$ according to \cref{eq:adv_mask2}. \\
    Update $\Phi$ with the gradient of \cref{eq:adv1} and learning rate of mask generator. \\
    // Update parameters of GraphMAE $\Theta$ \\
    Compute masking probability $prob$ according to \cref{eq:prob} and (\ref{eq:adv_weight}), and generate mask $\boldsymbol{m}$ based on $prob$ according to \cref{eq:adv_mask2}. \\
    Update $\Theta$ with the gradient of \cref{eq:adv3} and learning rate of GraphMAE. \\
}
Return representations learned by encoder of GraphMAE.
\end{algorithm}

\section{C. Details of Datasets}

We conduct experiments to compare the proposed \themodel with several baseline methods on fifteen public datasets in total, including nine node classification datasets (i.e., Cora, Citeseer~\cite{cora_citeseer}, Pubmed~\cite{pubmed}, Ogbn-arxiv~\cite{ogb}, PPI, Reddit, Corafull~\cite{corafull}, Flickr~\cite{flickr}, and WikiCS~\cite{WikiCS}), and six graph classification datasets (i.e., IMDB-B, IMDB-M, PROTEINS, COLLAB, MUTAG, and REDDIT-B~\cite{TUDataset}). 
\cref{tab:node-data} and \cref{tab:graph-data} provide the detailed  statistics about these datasets.

\section{D. Details of Baselines}

We compare with two types of SOTA graph self-supervised learning models as baselines: contrastive methods and generative methods.

\vspace{6pt}
\noindent\textbf{\textit{Contrastive methods:}}
\begin{itemize}
\item \textbf{Graph2vec}~\cite{Graph2vec} proposes a graph-level unsupervised representation learning technique based on subgraph extraction and negative sampling.
\item \textbf{DGI}~\cite{DGI} maximizes mutual information between patch representations and corresponding graph-level summaries to learn node representations.
\item \textbf{MVGRL}~\cite{MVGRL} introduces a self-supervised approach for learning node and graph level representations by contrasting structural views of graphs.
\item \textbf{GRACE}~\cite{GRACE} generates two graph views by corruption and learns node representations by maximizing the agreement of node representations in these two views.

\item \textbf{InfoGraph}~\cite{InfoGraph} maximizes the mutual information between the graph-level representation and the representations of substructures of different scales.
\item \textbf{GraphCL}~\cite{GraphCL} designs four types of graph augmentations to incorporate various priors and studies the impact of various combinations of graph augmentations in graph contrastive learning.
\item \textbf{GCC}~\cite{GCC} discriminates subgraph instancess in and across graphs and leverages contrastive learning to learn the intrinsic and transferable representations.
\item \textbf{JOAO}~\cite{JOAO} proposes a unified optimization framework to automatically, adaptively and dynamically select data augmentations when performing GraphCL.

\item \textbf{InfoGCL}~\cite{InfoGCL} studies how graph information is transformed and transferred and proposes an information-aware graph contrastive learning framework.
\item \textbf{CCA-SSG}~\cite{CCA-SSG} optimizes an innovative feature-level objective inspired by canonical correlation analysis, unlike traditional instance-level discrimination.
\item \textbf{BGRL}~\cite{BGRL} uses only simple augmentations and alleviates the need for contrasting with negative examples, and is thus scalable by design.
\end{itemize}

\noindent\textbf{\textit{Generative methods:}}
\begin{itemize}
\item \textbf{GraphMAE}~\cite{GraphMAE} explores generative self-supervised learning in graphs and proposes a simple yet effective masked graph autoencoder model.
\item \textbf{MaskGAE}~\cite{MaskGAE} adopts masked graph modeling (MGM) as a principled pretext task: masking a portion of edges and reconstructing the missing part with partially visible, unmasked graph structure.
\item \textbf{SeeGera}~\cite{SeeGera} enhances the family of self-supervised variational graph autoencoder on graph representation learning in a variety of downstream tasks.
\end{itemize}

\section{E. Experimental Settings}

We follow the evaluation protocol in GraphMAE~\cite{GraphMAE}. 
For node classification, we report the mean accuracy on the test nodes through 20 random initialization. For graph classification, we report the mean 10-fold cross-validation accuracy with standard deviation after 5 runs.
In each task, we follow exactly the same experimental procedure, e.g., data splits, evaluation protocol, and the standard settings

\section{F. Implementation Details}
\subsection{F.1. Hardware and Software.}
Our experiments are conducted on Linux servers equipped with an AMD CPU EPYC 7742 (256) @ 2.250GHz, 256GB RAM and NVIDIA 3090 GPUs. Our model is implemented in PyTorch version 1.11.0, DGL version 1.0.0 (https://www.dgl.ai/) with CUDA version 11.3, scikit-learn version 1.0.2 and Python 3.9.5. Our code is avalible at: \url{https://github.com/AzureLeon1/AUG-MAE}.

\subsection{F.2. Model Configuration.}
Adam is adopted as the optimizer. In node classification experiments, we adopt GAT as backbone of mask generator, encoder, and decoder. The learning rate of GraphMAE is set to 0.001, while the learning rate of mask generator is set to 0.0001.
In graph classification experiments, we adopt GIN as backbone of mask generator, encoder, and decoder. The learning rate of GraphMAE is set to 0.00015, while the learning rate of mask generator is set to 0.001.
We tune the weight of ratio of regularizer (i.e., $\lambda_1$) in \{1e-5, 1e-4, 1e-3, 1e-2, 1e-1, 1\}, tune the weight of uniformity regularizer (i.e., $\lambda_2$) in \{1e-5, 5e-5, 1e-4, 5e-4, 1e-3, 1e-2, 1e-1, 1, 10\}, tune $\alpha_0, \alpha_1$ in [0.0, 1.0], and tune $\eta$ in [0.5, 1.5] on each dataset. For baselines, we follow their recommended settings.

\end{document}